\pdfoutput=1

\documentclass[11pt]{article}

\usepackage[final]{coling}

\usepackage{times}
\usepackage{latexsym}

\usepackage[T1]{fontenc}

\usepackage[utf8]{inputenc}

\usepackage{microtype}

\usepackage{inconsolata}

\usepackage{graphicx}

\usepackage{booktabs}       
\usepackage{amsfonts}       
\usepackage{nicefrac}       
\usepackage{xcolor}         
\usepackage{wrapfig}
\usepackage{nicematrix}
\usepackage{mathtools}
\usepackage{multirow}
\usepackage{makecell}
\usepackage{array}
\usepackage{subcaption}
\usepackage{tabularx,colortbl}
\usepackage{enumitem}
\usepackage{commath}
\usepackage{soul}
\usepackage{amsmath}
\usepackage{float}
\usepackage{listings}
\usepackage{tcolorbox}
\usepackage{xspace}
\usepackage{cleveref}
\usepackage{multirow}
\usepackage{newfloat}
\usepackage{tablefootnote}
\usepackage[flushleft]{threeparttable}
\usepackage{seqsplit}
\usepackage{placeins}
\usepackage{amssymb, algorithm, algpseudocode}
\usepackage{MnSymbol} 

\usepackage{xcolor}
\usepackage{pifont}

\usepackage{hyperref}
\usepackage{url}

\usepackage{multicol}
\usepackage{fancyvrb}

\usepackage{amsmath,amsfonts,bm}

\def\eqref#1{equation~\ref{#1}}

\def\1{\bm{1}}

\DeclareMathAlphabet{\mathsfit}{\encodingdefault}{\sfdefault}{m}{sl}
\SetMathAlphabet{\mathsfit}{bold}{\encodingdefault}{\sfdefault}{bx}{n}

\newcommand{\subsec}[1]{\noindent\textbf{#1}~~}
\definecolor{ForestGreen}{RGB}{34, 139, 34}
\definecolor{TealBlue}{RGB}{0, 204, 204}

\crefname{section}{Sec.}{Secs.}
\crefname{table}{Table}{Tables}
\crefname{figure}{Fig.}{Figs.}

\definecolor{myGray}{rgb}{0.94,0.94,0.94}
\definecolor{magenta(dye)}{rgb}{0.79, 0.08, 0.48}
\definecolor{mycitecolor}{rgb}{0,0.08,0.75} %
\definecolor{arrow_blue}{rgb}{0,0.44,0.75} 
\definecolor{codegray}{rgb}{0.5,0.5,0.5}
\definecolor{codepurple}{rgb}{0.58,0,0.82}
\definecolor{backcolour}{rgb}{0.95,0.95,0.92}

\definecolor{darkmagenta}{RGB}{165, 0, 64}
\definecolor{darkcyan}{RGB}{0, 110, 175}

\definecolor{jsonKey}{rgb}{0.7, 0, 0} 
\definecolor{params}{rgb}{0, 0, 0.7} 

\newcommand{\eg}{e.g.\xspace}
\newcommand{\ie}{i.e.\xspace}
\newcommand{\class}[1]{{\texttt{#1}}}
\newcommand{\peeb}{PEEB\xspace}

\newcommand{\birdsoup}{Bird-11K\xspace}

\newcommand{\owlvitBase}{OWL-ViT$_{\text{\ensuremath{\mathsf{B/32}}}}$\xspace}

\newcommand{\cub}{CUB\xspace}
\newcommand{\nabirds}{NABirds\xspace}

\newif\ifcomments

\commentstrue

\ifcomments
\newcommand{\comments}[1]{#1}
\else
\newcommand{\comments}[1]{}
\fi

\soulregister{\ref}{7}
\soulregister{\cref}{7}
\soulregister{\Cref}{7}
\soulregister{\citep}{7}
\soulregister{\owlvitBase}{7}
\soulregister{\birdsoup}{7}
\soulregister{\cub}{7}
\soulregister{\nabirds}{7}
\soulregister{\subsec}{7}
\soulregister{\subsection}{7}
\soulregister{\peeb}{7}
\soulregister{\textbf}{7}
\soulregister{\ie}{7}

\lstdefinestyle{nohighlight}{
    basicstyle=\footnotesize\ttfamily, 
    breaklines=true,                    
    breakatwhitespace=true,            
    tabsize=2,                          
    xleftmargin=0pt,                    
    xrightmargin=0pt,                   
    frame=None                        
}

\makeatletter
\newcommand{\printfnsymbol}[1]{%
  \textsuperscript{\@fnsymbol{#1}}%
}
\makeatother

\title{SlimLM: An Efficient Small Language Model for \\ On-Device Document Assistance}

\author{
    Thang M. Pham$^\dagger$ \\
    \texttt{thangpham@auburn.edu} \\
    \And
    Phat T. Nguyen$^\ddagger$\\
    \texttt{pnguyen340@gatech.edu} \\
    \And
    Seunghyun Yoon$^\mathsection$ \\
    \texttt{syoon@adobe.com} \\
    \AND
    Viet Dac Lai$^\mathsection$ \\
    \texttt{daclai@adobe.com} \\
    \And
    Franck Dernoncourt$^\mathsection$ \\
    \texttt{franck.dernoncourt@gmail.com} \\
    \And
    Trung Bui$^\mathsection$ \\
    \texttt{bui@adobe.com} \\
    \AND
    $^\dagger$\textnormal{Auburn University} ~~~~~ $^\ddagger$\textnormal{Georgia Tech} ~~~~~ $^\mathsection$\textnormal{Adobe Research}
}

\begin{document}
\maketitle
\begin{abstract}
While small language models (SLMs) show promises for mobile deployment, their real-world performance and applications on smartphones remains underexplored. 
We present SlimLM, a series of SLMs optimized for document assistance tasks on mobile devices. 
Through extensive experiments on a Samsung Galaxy S24, we identify the optimal trade-offs between model size (ranging from 125M to 7B parameters), context length, and inference time for efficient on-device processing. 
SlimLM is pre-trained on SlimPajama-627B and fine-tuned on DocAssist, our constructed dataset for summarization, question answering and suggestion tasks. 
Our smallest model demonstrates efficient performance on S24, while larger variants offer enhanced capabilities within mobile constraints.
We evaluate SlimLM against existing SLMs, showing comparable or superior performance and offering a benchmark for future research in on-device language models.
We also provide a research demo, offering practical insights into SLM deployment. 
Our findings provide valuable insights and illuminate the capabilities of running advanced language models on high-end smartphones, potentially reducing server costs and enhancing privacy through on-device processing.
\end{abstract}

\section{Introduction}

The evolution of language models is diverging along two paths: large language models (LLMs) pushing the boundaries of artificial general intelligence in data centers \cite{chowdhery2022palm,gpt4,team2023gemini,touvron2023llama,touvron2023llama2,qwen_1,qwen_2.5}, and small language models (SLMs) designed for resource-efficient deployment on edge devices like smartphones \cite{mobilellama,mobillama,zhang2024tinyllama,liu2024mobilellm}. 
While LLMs have attracted significant attention, the practical implementation and performance of SLMs on real mobile devices remain understudied, despite their growing importance in consumer technology.

Recent developments, such as Qwen-2 \cite{qwen_2}, SmolLM \cite{smollm}, Gemini Nano \cite{reid2024gemini}, Apple Intelligence \cite{apple_intelligence} or LLaMA-3.2 \cite{meta_llama32} underscore the increasing relevance of SLMs in mobile applications. 
However, a comprehensive understanding of how these models perform on high-end smartphones is lacking.
Unlike previous works that primarily focus on developing smaller models without extensive real-device testing \cite{mobilellama,mobillama,zhang2024tinyllama,liu2024mobilellm}, our approach aims to bridge that gap by presenting an in-depth study of SLM development and deployment on a Samsung Galaxy S24 (also known as S24), focusing on three document assistance tasks: summarization (\class{SUMM}), question suggestion (\class{QS}), and question answering (\class{QA}).
By enabling efficient on-device document processing, our approach has the potential to significantly reduce server costs associated with API calls to cloud-based services, while enhancing user privacy.

We address critical questions about optimal model size, maximum context length, inference latency, memory constraints, and performance trade-offs on mobile devices.
To answer these questions, we introduce SlimLM, a series of small language models specifically designed and optimized for mobile deployment.
SlimLM is pretrained on the SlimPajama-627B \cite{cerebras2023slimpajama} and finetuned on DocAssist, our specialized dataset constructed based on $\sim$83K documents for document assistance.
Our models range from 125M to 1B parameters, allowing us to explore the full spectrum of what is possible on current mobile hardware.

Our results show that SlimLM models perform comparably or even better than existing SLMs of similar sizes across standard metrics such as BLEU \cite{papineni-etal-2002-bleu}, ROUGE \cite{lin-2004-rouge}, Semantic Textual Similarity (STS), Self-BLEU \cite{zhu2018selfbleu} for text diversity and GEval \cite{liu2023geval}. 
The smallest model SlimLM-125M demonstrates efficient performance on S24, making it suitable for widespread deployment.
Larger variants, up to 1B parameters, offer enhanced capabilities while still operating within mobile constraints.
To demonstrate real-world applicability, we develop a research demo showcasing SlimLM's document assistance capabilities (\cref{sec:use_case}).

Our key contributions are:
\begin{enumerate}
    \item We identify the sweet spot between model size, inference time, as well as the longest context length that can be efficiently processed on the latest Samsung device S24 (\cref{sec:sweet_spot}).
    \item We construct DocAssist, a specialized dataset for finetuning models on three critical document assistance tasks (\cref{sec:docassist}).
    \item We propose a set of small language models pretrained on SlimPajama with 627B tokens and finetuned on the DocAssist dataset (\cref{sec:slimlm}).
    \item SlimLM outperforms or performs comparably to existing SLMs of similar sizes while handling a maximum of 800 context tokens (\cref{sec:result}).
\end{enumerate}

\section{Approach}
\label{sec:approach}

To develop and deploy an efficient model for document assistance tasks on mobile devices, we propose a 3-step approach: (1) Determine an ideal model size that can handle sufficiently long context inputs in reasonable time; (2) Construct a dataset for instruction-finetuning models to enhance their document assistance capabilities; and (3) Train and fine-tune SlimLM, a series of models from scratch to perform document assistance tasks while running efficiently on mobile devices. 

\subsection{Sweet Spot: Model Size, Context Length and Inference Time}
\label{sec:sweet_spot}

Finding the sweet spot between model size, context length and inference time is important because larger models may take much time to handle and memory for being loaded so it cannot handle long context despite higher performance. Similarly, smaller models can handle longer contexts in a shorter time but it remains unknown how much their performance degrades.

\paragraph{Model Selection and Deployment} We select a list of state-of-the-art (SoTA) models ranging from 125M to 8B parameters as those larger than 8B are very challenging to be deployed even after quantization \cite{murthy2024mobileaibench}.
For quantization and deployment, we use the MLC-LLM framework \cite{mlc-llm} as it supports a wide range of SoTA models and GPU usage on mobile devices.
All models are quantized in 4-bit using the group quantization method with a group size of 32.

\paragraph{Context-length Selection} As document assistance tasks require handling long context inputs, we conduct experiments with different context lengths $L$ up to 1,000 tokens to measure the models' efficiency such as input token per second (ITPS), output token per second (OTPS), time to first token (TTFT) and total runtime in seconds.
A document is tokenized and the tokens are divided into $N = 5$ chunks, each chunk has a maximum of $\frac{max(L)}{N} = 200$ tokens.
We prepare one ($L = 200$), two ($L = 400$) and up to five chunks as context inputs to the models for summarizing.

\paragraph{Experiment}
We first start by asking five different short questions (less than 12 tokens) \eg ``Who was the first president of USA'' (\cref{tab:sweet_spot_no_context}) and measure their efficiency metrics to compute the average (\cref{tab:sweet_spot}a).
Next, we gradually add more input contexts \ie chunks extracted from five different documents as described along with different requests (\cref{tab:sweet_spot_with_context}) to prompt the models for the summarization task and record the average results (\cref{tab:sweet_spot}b--e).

\paragraph{Results}
\cref{tab:sweet_spot} presents a clear trade-off between model size and speed, with smaller models like SmolLM or Qwen2 showing higher inference speeds (IPTS, TTFT) but potentially lower accuracy compared to larger models (\eg Gemma-2, Phi-3.5, Mistral or Llama-3.1).
As input length increases, most models experience decreased inference speeds, highlighting the impact of prompt size on efficiency. 
When the input context reaches approximately 1,000 tokens (5 chunks), smaller models (\eg SmolLM, Qwen2) struggle to complete multiple experimental runs, while larger models face memory constraints on these long inputs. 
Mid-sized models like Qwen2-0.5B-Instruct often strike a balance between speed, accuracy, and input handling capacity, potentially offering the best compromise for practical applications within certain input length constraints.

\begin{table}[ht]
\centering
\vspace*{\fill}
\resizebox{0.489\textwidth}{!}{%
\begin{tabular}{|l|r|r|r|r|}
\hline
\rowcolor[HTML]{A9A9A9} 
Model & ITPS (t/s) & OTPS (t/s) & TTFT (s) & Runtime (s) \\ 
\rowcolor[HTML]{D3D3D3} 
\multicolumn{5}{|c|}{(a) Prompt: ``Who was the first president of USA?''} \\ \hline
SmolLM-135M-Instruct & 68.48 & 59.72 & 0.46 & 1.42 \\ \hline
SmolLM-360M-Instruct & 27.56 & 56.68 & 0.85 & 3.71 \\ \hline
Qwen2-0.5B-Instruct & 23.84 & 51.78 & 1.90 & 2.38 \\ \hline
Qwen2-1.5B-Instruct & 3.42 & 17.12 & 13.01 & 14.39 \\ \hline
Gemma-2-2b-it & 1.82 & 18.64 & 10.56 & 13.52 \\ \hline
Phi-3-mini-4k-instruct & 0.86 & 14.78 & 39.81 & 48.29 \\ \hline
Phi-3.5-mini-instruct & 0.88 & 15.60 & 39.90 & 47.49 \\ \hline
Mistral-7B-Instruct-v0.3 & 0.44 & 9.36 & 127.60 & 135.12 \\ \hline
Llama-3.1-8B-Instruct & 0.10 & 2.20 & 261.65 & 269.99 \\ \hline
\rowcolor[HTML]{D3D3D3} 
\multicolumn{5}{|c|}{(b) Prompt: 1 chunk $\sim$ 200 tokens (157 words)} \\ \hline
SmolLM-135M-Instruct & 167.80 & 60.80 & 1.91 & 4.22 \\ \hline
SmolLM-360M-Instruct & 28.42 & 36.12 & 10.62 & 16.82 \\ \hline
Qwen2-0.5B-Instruct & 23.02 & 39.42 & 13.15 & 14.96 \\ \hline
Qwen2-1.5B-Instruct & 3.86 & 14.70 & 78.78 & 86.14 \\ \hline
Gemma-2-2b-it & 2.20 & 11.68 & 122.06 & 141.15 \\ \hline
Phi-3-mini-4k-instruct & 1.05 & 12.68 & 327.09 & 339.87 \\ \hline
\rowcolor[HTML]{D3D3D3} 
\multicolumn{5}{|c|}{(c) Prompt: 2 chunks $\sim$ 400 tokens (269 words)} \\ \hline
SmolLM-135M-Instruct & 130.66 & 40.42 & 4.84 & 8.14 \\ \hline
SmolLM-360M-Instruct & 23.28 & 27.90 & 30.40 & 41.07 \\ \hline
Qwen2-0.5B-Instruct & 18.62 & 24.72 & 29.49 & 38.36 \\ \hline
\rowcolor[HTML]{D3D3D3} 
\multicolumn{5}{|c|}{(d) Prompt: 3 chunks $\sim$ 600 tokens (368 words)} \\ \hline
SmolLM-135M-Instruct & 174.10 & 45.70 & 4.89 & 8.26 \\ \hline
SmolLM-360M-Instruct & 31.50 & 33.94 & 27.16 & 33.52 \\ \hline
Qwen2-0.5B-Instruct & 20.53 & 25.04 & 37.94 & 47.05 \\ \hline
\rowcolor[HTML]{D3D3D3} 
\multicolumn{5}{|c|}{(e) Prompt: 4 chunks $\sim$ 800 tokens (529 words)} \\ \hline
SmolLM-135M-Instruct & 134.66 & 32.96 & 8.47 & 11.83 \\ \hline
SmolLM-360M-Instruct & 23.60 & 25.52 & 48.06 & 58.15 \\ \hline
Qwen2-0.5B-Instruct & 19.74 & 19.52 & 54.90 & 66.65 \\ \hline
\end{tabular}
}
\vspace*{\fill}
\caption{Performance comparison of language models across varying input lengths ranging from single questions to chunks of around 800 tokens. 
Smaller models demonstrate higher efficiency but potentially lower accuracy, while larger models generally exhibit slower inference speeds but better handling of longer inputs.
}
\label{tab:sweet_spot}
\end{table}



\subsection{Document Assistance Dataset}
\label{sec:docassist}

While smaller models offer faster inference speeds, they often have limited document-handling capabilities. 
To address this, we develop DocAssist, a specialized dataset designed for fine-tuning these models to enhance their ability to process and assist with longer documents.

\subsubsection{Data Collection}

We utilize our proprietary tools to compile a diverse collection of documents, primarily consisting of illustrations, presentation slides, and spreadsheets. 
This dataset also includes machine-generated documents to ensure a comprehensive representation of various document types.
We extract the document contents and prepare them for pre-processing to ensure the data is suitable for model fine-tuning.

\paragraph{Pre-processing}
We employ Tiktoken \cite{tiktoken} to tokenize the documents. 
Each document is segmented into 5 chunks, with each chunk containing a maximum of 200 tokens. 
This segmentation ensures that the maximum number of tokens per document after pre-processing is 1,000. 
Consequently, documents with fewer than 1,000 tokens remain unaltered, while longer documents are truncated. 
\cref{tab:token_dist_comparison} presents the statistical analysis of token distribution per document, including the mean, standard deviation, and range of token counts, both before and after pre-processing.

\begin{table}[ht]
\centering
\resizebox{0.45\textwidth}{!}{%
\begin{tabular}{|l|c|c|}
\hline
\rowcolor[HTML]{A9A9A9} 
Processing Stage & Mean $\pm$ STD & Token Range \\ \hline
Pre-processing  & 8,635 $\pm$ 24,235 & 1 -- 1,675,639 \\ \hline
Post-processing &   879 $\pm$    252 & 1 -- 1,000 \\ \hline
\end{tabular}
}
\caption{Statistical comparison of token distribution per document before and after pre-processing the documents. The table shows the mean $\pm$ standard deviation and the range of token counts for each processing stage.}
\label{tab:token_dist_comparison}
\end{table}

\subsubsection{Data Annotation}


We propose an approach for annotating documents using a commercial LLM to generate comprehensive annotations for three key tasks in DocAssist: \class{SUMM}, \class{QS}, and \class{QA}. 
For each document, our method produces five distinct examples: one summary, one set of three suggested questions, and three question-answer pairs.

\paragraph{Prompt Design}
Our annotation process employs a carefully designed prompt (\cref{tab:anno_prompt}) that instructs the model to perform these tasks sequentially. 
The prompt is applied to each processed document, replacing the \texttt{\textcolor{blue}{\{\{document\}\}}} placeholder with the actual content. 
The annotation prompt elicits a JSON response containing a document summary, three suggested questions, and their corresponding answers.
To ensure high-quality and diverse annotations, we incorporate task-specific requirements:
\begin{enumerate}
    \item \texttt{\textcolor{violet}{\{\{summ\_req\}\}}}: to produce concise, informative overviews that capture the document's essence, enabling models to recognize and respond to requests for document overview.
    \item \texttt{\textcolor{orange}{\{\{suggestion\_req\}\}}}: to generate diverse, relevant questions probing different aspects of the document's content, allowing models to assist users seeking guidance on what to ask about a document or topic.
    \item \texttt{\textcolor{teal}{\{\{qa\_req\}\}}}: to provide accurate, contextually appropriate answers to document-specific questions, training models to recognize and respond to user queries for specific information or explanations from the document.
\end{enumerate}

Our approach serves several crucial functions: it facilitates intent classification training, enables task-specific response generation, enhances contextual understanding, ensures versatility in document handling, and maintains quality control in annotations. 
By leveraging the capabilities of the commercial LLM, we aim to generate high-quality annotations that capture the nuances and complexities of the documents.
The in-context examples and detailed requirements are provided in \cref{tab:anno_prompt_full,tab:do_prompt,tab:qa_prompt,tab:sq_prompt}.

\begin{table}[ht]
\centering
\begin{tcolorbox}[
    colback=white,
    colframe=black,
    arc=0mm,
    boxrule=0.5pt,
    width=0.48\textwidth
]
\tiny

You will be given a document. Your task is to provide a summary of the document, suggest relevant questions, and then answer those questions.

\textbf{Task Requirements:}
\begin{enumerate}[noitemsep,leftmargin=*]
    \item Summarization: \texttt{\textcolor{violet}{\{\{summ\_req\}\}}}
    \item Question Suggestion: \texttt{\textcolor{orange}{\{\{suggestion\_req\}\}}}
    \item Question Answering: \texttt{\textcolor{teal}{\{\{qa\_req\}\}}}
\end{enumerate}

Format your response in JSON as shown in the examples below.
\begin{tcolorbox}[colback=gray!10,colframe=gray!10,left=1pt,right=1pt,top=1pt,bottom=1pt]
\tiny
\begin{verbatim}
{"tasks": {
 "summarization": "Your summary here...",
 "question_suggestion": [...],
 "question_answering": [...]}}
\end{verbatim}
\end{tcolorbox}

\textbf{Examples:} \\

[Two in-context examples here]

\textbf{\\DOCUMENT CONTEXT} (may be truncated)

\texttt{\textcolor{blue}{\{\{document\}\}}} \\

\textbf{RESPONSE}

\end{tcolorbox}

\caption{A prompt designed to annotate data for three tasks given a document in DocAssist: \class{SUMM}, \class{QS} and \class{QA}.
\texttt{\textcolor{blue}{\{\{document\}\}}} is replaced with each pre-processed document.
Please see the complete prompt with in-context examples and requirements for each task \texttt{\textcolor{violet}{\{\{summ\_req\}\}}}, \texttt{\textcolor{orange}{\{\{suggestion\_req\}\}}} and \texttt{\textcolor{teal}{\{\{qa\_req\}\}}} in \cref{tab:anno_prompt_full,tab:do_prompt,tab:qa_prompt,tab:sq_prompt}, respectively.
}
\label{tab:anno_prompt}
\end{table}

\paragraph{Result}
\cref{tab:token_usage_stats} provides insight into the token usage statistics for the commercial LLM in annotating the documents.
The relatively low standard deviation in completion tokens suggests consistent-length responses across different documents, which is desirable for maintaining annotation quality and consistency.
The annotation process yields $\sim$414K examples for DocAssist. 
Of these, $\sim$2K examples were randomly selected for the test set, with the remaining examples allocated to the training set.

\begin{table}[ht]
\centering
\resizebox{0.48\textwidth}{!}{%
\begin{tabular}{|l|c|c|}
\hline
\rowcolor[HTML]{A9A9A9} 
Token Type & Mean $\pm$ STD & Token Range \\ \hline
Prompt Tokens    & 2,126.04 $\pm$ 260.81 & 1,273 -- 2,617 \\ \hline
Completion Tokens &  169.07 $\pm$  17.61 &   107 --   312 \\ \hline
\end{tabular}
}
\caption{Token usage statistics for the commercial LLM in annotating the documents.}
\label{tab:token_usage_stats}
\end{table}


\subsection{Slim Language Model}
\label{sec:slimlm}

SlimLM is based on the MPT (Mosaic Pre-trained Transformer) architecture by \citealp{MosaicML2023Introducing} with specific modifications to optimize for document assistance tasks.
Specifically, we opt not to use the ALiBi \cite{alibi} positioning method as document assistance tasks primarily deal with fixed-length inputs and outputs.
Unlike the original MPT, SlimLM incorporates biases in its layers to enhance the model's flexibility in capturing and representing document-specific nuances.
Biases can help the model learn task-specific offsets, potentially improving its ability to distinguish between \class{SUMM}, \class{QS}, and \class{QA} tasks.
Based on the sweet-spot findings (\cref{sec:sweet_spot}), we create and train a range of models from 125M to 1B parameters by adjusting the number of layers and heads.

\subsubsection{Pre-training}

We pre-trained SlimLM on the SlimPajama dataset \cite{cerebras2023slimpajama}, comprising 627B tokens. 
The pre-training objective follows the standard autoregressive language modeling approach, where the model learns to predict the next token in the sequence. 
The loss function for pre-training can be expressed as:
\begin{equation}
    L_{pt} = -\sum_{i=1}^{n} \log P(x_i | x_{<i})
\end{equation}

where $x_i$ represents the $i^{th}$ token in the input sequence, $x_{<i}$ denotes all tokens preceding $x_i$, and $n$ is the length of the sequence.

\subsubsection{Fine-tuning}

Following pre-training, we fine-tuned the models on the training set of DocAssist that comprises $\sim$412K examples to enhance document assistance capabilities by teaching them to handle specific tasks based on user requests. 
The process instructs the model to first identify the appropriate task from the user's input and then generate a response that match the quality of the commercial LLM for the identified task.
The fine-tuning loss function is also an autoregressive objective, defined as:
\begin{equation}
    L_{ft} = -\sum_{i=1}^{m} \log P(y_i | y_{<i}, x)
\end{equation}

where $x$ is the input sequence (system prompt, document and user request), $y_i$ is the $i^{th}$ token in the target response generated by the commercial LLM, $y_{<i}$ denotes all tokens preceding $y_i$ in the target response $m$ is the length of the target response.

\section{Experiments and Results}
\label{sec:result}

\subsection{Experiment Setup}

We pre-train SlimLM from scratch on the SlimPajama dataset using 128-256 A100/H100 GPUs using Lion optimizer \cite{chen2023symbolicdiscoveryoptimizationalgorithms} with different learning rates (LRs), global batch size, and number of trained tokens. 
All models are fine-tuned on DocAssist using 8 A100 GPUs using AdamW optimizer \cite{loshchilov2017decoupled} with the same LR of 5e-6 and global batch size of 48.
The models' configurations and hyperparameters are in \cref{tab:slimlm_models}.

\subsubsection{Baselines}
Our selection is based on the sweet-spot results that demonstrate a clear trade-off between model size, speed, and context length.
Specifically, we compare with the following models: SmolLM-135M-Instruct, SmolLM-360M-Instruct \cite{smollm}, Qwen2-0.5B-Instruct and Qwen2-1.5B-Instruct \cite{qwen_2}.
These models represent SoTA performance at their respective sizes, making them strong baselines for comparison.

\subsubsection{Evaluation Metrics}
We employ a diverse set of metrics to evaluate models' performance across the DocAssist tasks.
For Intent Detection, we use Accuracy to measure classification precision. 
\class{SUM}, \class{QS}, and \class{QA} tasks are evaluated using BLEU \cite{papineni-etal-2002-bleu}, ROUGE \cite{lin-2004-rouge}, and Semantic Textual Similarity (STS) scores, which assess the quality, overlap, and semantic similarity of generated outputs compared to references. 
GEval \cite{liu2023geval} provide a comprehensive quality assessment with human alignment for \class{SUMM} and \class{QA}\footnote{We adjust GEval prompts originally designed for summarization task accordingly for the evaluation of QA task.} outputs.
While other metrics have scores in the range [0, 1], GEval scores range from 1 to 4.5. To ensure consistency across metrics, we rescale GEval scores to the same interval.
Additionally, we use Self-BLEU for Text Diversity \cite{zhu2018selfbleu} for \class{QS} to ensure varied outputs.

\subsection{Results}
Before finetuning, all models cannot perform document assistance tasks or detect user intents. After finetuning, most models achieve perfect accuracy, with the lowest score being 99.86\% from SmolLM-360M-Instruct (\cref{tab:intent_cls}).
\cref{tab:main} demonstrates the effectiveness of our SlimLM models compared to the baselines across the three DocAssist tasks. 
Specifically, SlimLM models consistently outperform or match the performance of similar-sized counterparts, indicating the efficiency of our architecture.
SlimLM-125M surpasses SmolLM-135M-Instruct, while both SlimLM-270M and SlimLM-350M outperform SmolLM-360M-Instruct. 
Notably, SlimLM-450M and SlimLM-760M achieve comparable results to Qwen2-0.5B-Instruct, despite the latter being pre-trained and fine-tuned on a substantially larger dataset.
Detailed results for each task are presented in the appendix (\cref{tab:main_summ,tab:main_qa,tab:main_qs}).

As model size increases (\cref{tab:main}), we observe consistent improvement across all metrics, suggesting good scalability.
Our largest model, SlimLM-1B, approaches the performance of the much larger model Qwen2-1.5B-Instruct, highlighting the potential for SlimLM to achieve competitive results with reduced computational requirements.
While the commercial LLM still leads in overall performance, our SlimLM models offer a range of efficient options for various computational constraints and privacy concerns in document assistance tasks.


\section{Use Case}
\label{sec:use_case}


\begin{table*}[ht]
\vspace*{\fill}
\centering
\resizebox{0.8\textwidth}{!}{%
\begin{tabular}{|l|c|c|c|c|c|c|c|}
\hline
\rowcolor[HTML]{A9A9A9} 
Model & BLEU $\uparrow$ & ROUGE-1 $\uparrow$ & ROUGE-2 $\uparrow$ & ROUGE-L $\uparrow$ & STS Score $\uparrow$ & GEval $\uparrow$ & Average \\
\hline
The commercial LLM & 1.00 & 1.00 & 1.00 & 1.00 & 1.00 & 0.88 & 0.9795 \\
\hline
SmolLM-135M-Instruct & \cellcolor{green!25}0.10 & \cellcolor{green!25}0.37 & \cellcolor{green!25}0.17 & \cellcolor{green!25}0.34 & \cellcolor{green!25}0.64 & \cellcolor{green!25}0.60 & \cellcolor{green!25}0.3694 \\
SmolLM-360M-Instruct & \cellcolor{green!25}0.14 & \cellcolor{green!25}0.42 & \cellcolor{green!25}0.21 & \cellcolor{green!25}0.38 & \cellcolor{green!25}0.68 & \cellcolor{green!25}0.69 & \cellcolor{green!25}0.4202 \\
Qwen2-0.5B-Instruct & 0.21 & 0.49 & 0.28 & 0.45 & 0.74 & 0.79 & 0.4934 \\
Qwen2-1.5B-Instruct & 0.26 & 0.53 & 0.33 & 0.50 & 0.77 & 0.84 & 0.5396 \\
LLaMA-3.2-1B-Instruct & 0.26 & 0.53 & 0.33 & 0.50 & 0.77 & 0.86 & \textbf{0.5442} \\
\hline
\rowcolor[HTML]{E6E6E6}
\multicolumn{8}{|l|}{\textbf{Slim Language Models (ours)}} \\
\hline
SlimLM-125M$^{a}$ & \cellcolor{green!25}\textbf{0.14} & \cellcolor{green!25}\textbf{0.41} & \cellcolor{green!25}\textbf{0.21} & \cellcolor{green!25}\textbf{0.38} & \cellcolor{green!25}\textbf{0.66} & \cellcolor{green!25}\textbf{0.64} & \cellcolor{green!25}\textbf{0.4052} \\
SlimLM-270M & 0.17 & 0.45 & 0.24 & 0.42 & 0.71 & 0.72 & 0.4497 \\
SlimLM-350M$^{b}$ & \cellcolor{green!25}\textbf{0.18} & \cellcolor{green!25}\textbf{0.45} & \cellcolor{green!25}\textbf{0.25} & \cellcolor{green!25}\textbf{0.42} & \cellcolor{green!25}\textbf{0.71} & \cellcolor{green!25}\textbf{0.73} & \cellcolor{green!25}\textbf{0.4541} \\
SlimLM-450M$^{c}$ & 0.20 & 0.48 & 0.27 & 0.44 & 0.73 & 0.76 & 0.4806 \\
SlimLM-760M & 0.21 & 0.48 & 0.28 & 0.45 & 0.74 & 0.79 & 0.4911 \\
SlimLM-1B$^{d}$ & 0.23 & 0.51 & 0.31 & 0.48 & 0.76 & 0.81 & 0.5182 \\
\hline
\end{tabular}
}
\caption{\textbf{Comparison of model performance on average of three tasks: \class{SUMM}, \class{QS} and \class{QA}}. 
Green highlighting indicates superior performance of SlimLM models compared to similar-sized counterparts. 
Key comparisons: (a) SlimLM-125M outperforms SmolLM-135M-Instruct, (b) SlimLM-350M exceeds SmolLM-360M-Instruct, (c) SlimLM-450M is comparable to Qwen2-0.5B-Instruct, and (d) SlimLM-1B approaches Qwen2-1.5B-Instruct despite being smaller.
\cref{tab:main_summ,tab:main_qa,tab:main_qs} present detailed results for each task.}
\label{tab:main}
\vspace*{\fill}
\end{table*}

SlimLM can be deployed into the mobile apps, enabling local processing of documents.
This approach eliminates the need for external API calls, substantially reducing operational costs while enhancing user privacy by keeping document content on the device.

When a document is loaded, such as a legal contract, the app instantly generates a summary, suggests relevant questions, and provides quick answers to user queries, all without internet connectivity. 
This streamlined process allows professionals to grasp essential information rapidly and identify areas needing closer examination while maintaining document confidentiality and improving overall user experience.
Users can also interact with the document by chatting with the AI assistant. 


\section{Related Work}
\label{sec:related_work}

\subsection{Small and Large Language Models}

Large language models \cite{chowdhery2022palm,chung2022scaling,touvron2023llama,touvron2023llama2} have demonstrated impressive capabilities across various NLP tasks. 
However, their massive size limits practical deployment, especially on resource-constrained devices. 
This has spurred interest in small language models \cite{microsoft_phi_2,microsoft_phi_3_mini,bai2023qwen,gemma_2} that balance performance and efficiency.
While some approaches focus on compressing LLMs through techniques like knowledge distillation \cite{gu2023knowledge,zhang2024tinyllama}, our work aligns more closely with efforts to design and train efficient SLMs from scratch \cite{liu2024mobilellm,mobillama}. 
These approaches aim to achieve competitive performance with smaller model sizes and less training data.
Our SlimLM builds on these efforts by focusing specifically on optimizing SLMs for document processing tasks on mobile devices.

\subsection{SLMs for Mobile Devices}

Deploying language models on mobile devices presents unique challenges, including memory constraints, inference latency, and energy efficiency \cite{liu2024mobilellm,mobillama,chen2024octopus}. 
The growing importance of efficient on-device language models is further underscored by recent developments from major tech companies \cite{reid2024gemini,apple_intelligence,meta_llama32}. 
Our work extends this line of research by identifying the optimal balance between model size, context length, and performance specifically for real mobile devices \eg Samsung Galaxy S24. 
We focus on enhancing document assistance abilities by designing and training SlimLM (Sec.~\ref{sec:slimlm}) from scratch on SlimPajama and DocAssist (Sec.~\ref{sec:docassist}), advancing the SoTA in mobile-deployed language models for document processing applications.

\section{Conclusion}
\label{sec:conclusion}

In this work, we introduce SlimLM models optimized for document assistance tasks. 
We identify the optimal balance between model size, inference time, and maximum context length for efficient processing on real mobile devices. 
Our specialized DocAssist dataset, constructed from $\sim$83K documents, enabled fine-tuning of SlimLM for three critical document assistance tasks. 
SlimLM models, ranging from 125M to 1B parameters, demonstrate comparable or superior performance to existing SLMs of similar sizes across standard metrics, while efficiently handling up to 800 context tokens. 
To showcase real-world applicability, we develop a research demo featuring SlimLM's document assistance capabilities, paving the way for widespread deployment of efficient, on-device language models for enhanced user privacy and reduced server costs.

\bibliography{custom}

\clearpage

\appendix

\section{Appendix}
\label{sec:appendix}

\begin{table}[ht]
\centering
\resizebox{0.35\textwidth}{!}{%
\begin{tabular}{|l|r|}
\hline
\rowcolor[HTML]{A9A9A9}
Model & Accuracy (\%) \\
\hline
The commercial LLM & 100.00 \\
\hline
SmolLM-135M-Instruct & 99.86 \\
\hline
SmolLM-360M-Instruct & 99.81 \\
\hline
Qwen2-0.5B-Instruct & 100.00 \\
\hline
Qwen2-1.5B-Instruct & 100.00 \\
\hline
SlimLM-125M & 100.00 \\
\hline
SlimLM-270M & 100.00 \\
\hline
SlimLM-350M & 100.00 \\
\hline
SlimLM-450M & 100.00 \\
\hline
SlimLM-760M & 99.95 \\
\hline
SlimLM-1B & 99.90 \\
\hline
\end{tabular}
}
\caption{Intent Classification accuracy of various language models after fine-tuning on DocAssist dataset.}
\label{tab:intent_cls}
\end{table}

\FloatBarrier
\begin{table}[ht]
\centering
\begin{tcolorbox}[
    colback=white,
    colframe=black,
    arc=0mm,
    boxrule=0.5pt,
    width=0.5\textwidth
]
\small
Q1: Who was the first president of USA? \\
Q2: What is the capital city of France? \\
Q3: Who was the first person to walk on the moon? \\
Q4: What is the chemical symbol for gold? \\
Q5: In what year did World War II end?

\end{tcolorbox}

\caption{Fact-checking questions asked to measure a model's efficiency on real mobile devices.}
\label{tab:sweet_spot_no_context}
\end{table}
\FloatBarrier

\begin{table}[ht]
\centering
\begin{tcolorbox}[
    colback=white,
    colframe=black,
    arc=0mm,
    boxrule=0.5pt,
    width=0.5\textwidth
]
\small
R1. Please summarize the document excerpt(s) below: \\
R2. Kindly provide a concise overview of the following document excerpt(s): \\
R3. Briefly outline the main points from the passage(s) below: \\
R4. Highlight the key ideas from the following text sample(s): \\
R5. Capture the key points of the document snippet(s) provided:

\end{tcolorbox}

\caption{Summarizing requests used to measure a model's efficiency with different input contexts on real mobile devices.}
\label{tab:sweet_spot_with_context}
\end{table}

\begin{table}[ht]
\centering
\begin{tcolorbox}[
    colback=white,
    colframe=black,
    arc=0mm,
    boxrule=0.5pt,
    width=0.5\textwidth
]
\small

Summarize the main topic and key points of this document in one concise sentence. Ensure the summary gives a clear overview of the document's content without including minor details.

\end{tcolorbox}

\caption{
\texttt{\textcolor{violet}{\{\{summ\_req\}\}}}.
Instructional prompt designed to guide the commercial LLM how to summarize the document contents.}
\label{tab:do_prompt}
\end{table}

\begin{table}[ht]
\centering
\begin{tcolorbox}[
    colback=white,
    colframe=black,
    arc=0mm,
    boxrule=0.5pt,
    width=0.5\textwidth
]
\small

\textbf{Provide answers to the suggested questions, adhering to the following guidelines:} \\

a. Answer each question directly and completely based on the information in the document. \\
b. Provide specific details, explain your reasoning and, if applicable, cite relevant parts of the document. \\
c. Keep answers concise but informative, typically 1-3 sentences each. \\
d. If a question cannot be fully answered based solely on the document, state this clearly and provide the best possible answer with the available information. \\
e. Ensure that answers are accurate and directly related to the corresponding questions.

\end{tcolorbox}

\caption{
\texttt{\textcolor{teal}{\{\{qa\_req\}\}}}.
Instructional prompt designed to guide the commercial LLM how to answer questions for the Q/A task.}
\label{tab:qa_prompt}
\end{table}

\begin{table}[ht]
\centering
\begin{tcolorbox}[
    colback=white,
    colframe=black,
    arc=0mm,
    boxrule=0.5pt,
    width=0.5\textwidth
]
\small

Generate \textit{three insightful questions} so a user can explore and understand the document better and more quickly. \\

\textbf{When generating the questions, please consider the following:} \\
a. What questions am I interested in asking as a reader? \\
b. What questions does this document actually answer? \\

\textbf{Please make sure to adhere to the following specifications:} \\
a. Questions must be short and simple. \\
b. Each question must be less than 12 words. \\
c. You must not write questions that are too general. For example, ``what is this document about?'' or ``what is the purpose of this document'' are bad questions. \\
d. Questions must be specific to the document. For example, you should consider using entities and proper nouns that appear in the document to write your question, whenever possible. \\
e. Questions must have an answer based on the document I am reading. \\
f. Questions must be diverse, covering different parts of the document. \\ 
g. Please generate exactly 3 questions. \\

\end{tcolorbox}

\caption{
\texttt{\textcolor{orange}{\{\{suggestion\_req\}\}}}.
Instructional prompt designed to guide the commercial LLM how to generate suggested questions for a given document. The suggested questions aims to guide users what should be asked to understand the document.}
\label{tab:sq_prompt}
\end{table}

\begin{table*}[ht]
\centering
\begin{tcolorbox}[
    colback=white,
    colframe=black,
    arc=0mm,
    boxrule=0.5pt,
    width=\textwidth
]
\tiny

You will be given a document. Your task is to provide a summary of the document, suggest relevant questions, and then answer those questions.

\textbf{Task Requirements:}
\begin{enumerate}[noitemsep,leftmargin=*]
    \item Summarization: \texttt{\textcolor{violet}{\{\{summ\_req\}\}}}
    \item Question Suggestion: \texttt{\textcolor{orange}{\{\{suggestion\_req\}\}}}
    \item Question Answering: \texttt{\textcolor{teal}{\{\{qa\_req\}\}}}
\end{enumerate}

Format your response in JSON as shown in the examples below.
\begin{tcolorbox}[colback=gray!10,colframe=gray!10,left=1pt,right=1pt,top=1pt,bottom=1pt]
\tiny
\begin{verbatim}
{
  "tasks": {
    "summarization": "Your summary here...",
    "question_suggestion": ["Question 1?", "Question 2?", "Question 3?"],
    "question_answering": ["Answer to question 1.", "Answer to question 2.", "Answer to question 3."]
  }
}
\end{verbatim}
\end{tcolorbox}

\textbf{Examples:} \\

\textbf{DOCUMENT CONTEXT} (may be truncated)

\text{\#\#\#} Document Excerpt 1: English Language Arts, Grade 9 Inquiry question: How does an individual address injustice in society? Abstract This sample highlights my thinking and planning as it relates to a poetry lesson based on the redesigned BC Curriculum. I have made explicit connections to the First Peoples Principles of Learning, the Big Ideas, the Curricular Competencies and Content learning standards, Core Competencies, and. I have deliberately chosen activities that offer opportunities for multiple entry points, choice, and differentiation. I have also included some samples of student work. Through the exploration of text, discussion, and the activities of the lesson, students have opportunities to respond to the inquiry question with an informed and thoughtful voice. As students engage with text, collaborate with one another, create personal responses to text, and finally reflect on the process of their learning, they are doing so purposefully, considering the inquiry question. Student learning depends on considering prior knowledge, sharing and developing ideas, and revising and extending thinking. A(truncated due to length...) \text{\#\#\#} Document Excerpt 2: Understand text differently depending on their worldviews and perspectives. Texts are socially, culturally, and historically constructed. Questioning what we hear, read, and view contributes to our ability to be educated and engaged citizens. Curricular Competencies. Think critically, creatively, and reflectively to explore ideas within, between, and beyond texts. Construct meaningful personal connections between self, text, and world. Respond to text in personal, creative, and critical ways. Use writing and design processes to plan, develop, and create engaging and meaningful literary and informational texts for a variety of purposes and audiences Content Strategies and processes: o Writing processes Core Competencies The Learning Experience Exploring the concept of injustice Students are invited to bring to class examples of injustice in our community and/or in our world. This provides multiple entry points into the inquiry for the students. These examples provide opportunities for students to discuss the diverse ways in which injustice is evident/experienced in their(truncated due to length...) \\
... \\

\textbf{RESPONSE}

\begin{tcolorbox}[colback=gray!10,colframe=gray!10,left=1pt,right=1pt,top=1pt,bottom=1pt]
\tiny
\begin{verbatim}
"tasks": {
    "summarization": "The document outlines a Grade 9 poetry lesson focused on addressing injustice, using inquiry-based activities aligned with 
    the redesigned BC Curriculum and emphasizing critical thinking, personal connections, and creative expression.",
    "question_suggestion": [
        "What inquiry question is central to the lesson?",
        "How do students engage with the concept of injustice?",
        "What is the purpose of the empathy journal?"
    ],
    "question_answering": [
        "The central inquiry question is: How does an individual address injustice in society?",
        "Students engage with the concept of injustice by discussing real-world examples and exploring texts that highlight these issues.",
        "The purpose of the empathy journal is to help students transform negative messages into positive ones and reflect on the emotional impact of 
        those transformations."
    ]
}
\end{verbatim}
\end{tcolorbox}
 
\textbf{\\DOCUMENT CONTEXT} (may be truncated)

\text{\#\#\#} Document Excerpt 1: Demographic situation in Bulgaria in 2018: Number of population decreases and population ageing continue; Unbalanced territorial distribution of population goes deeper; Number of live births and crude birth rate decrease;  Number of deaths and crude death rate decrease; Infant mortality decreases; Number of marriages and number of divorces increase; Life expectancy at birth is unchanged. Population number and structures 1 As of 31 December 2018 the population of Bulgaria is 7 000 039 persons representing 1.4\% of the EU population. Compared to 2017, the country population decreases by 49 995 persons or by 0.7\%. Male population was 3 395 701 (48.5\%) and female - 3 604 338 (51.5\%) or to 1 000 males correspond 1 061 females. Males prevail among the population aged up to 53(truncated due to length...) \text{\#\#\#} Document Excerpt 2: the end of 2018, the number of population over working age is 1 732 thousand persons or 24.8\% and under working age are 1 067 thousand persons or 15.2\% of the country population. 1. Population under, at and over working age as of 31.12. Year Age Under working age Working age Over working age (Per cent) 1990 21.6 55.5 22.9 1995 19.1 56.6 24.3 2001 16.3 59.2 24.5 2005 14.8 62.4 22.8 2010 14.6 62.7 22.7 2011 14.3 61.7 24.0 2012 14.4 61.8 23.8 2013 14.6 (truncated due to length...) \\
... \\

\textbf{RESPONSE}
\begin{tcolorbox}[colback=gray!10,colframe=gray!10,left=1pt,right=1pt,top=1pt,bottom=1pt]
\tiny
\begin{verbatim}
"tasks": {
    "summarization": "In 2018, Bulgaria's population declined to 7 million, with notable demographic changes including decreased birth rates,
    increased divorces, and a high mortality rate compared to EU averages.",
    "question_suggestion": [
        "What was Bulgaria's population at the end of 2018?",
        "How did the birth and death rates change?",
        "Which municipality had the highest population in 2018?"
    ],
    "question_answering": [
        "At the end of 2018, Bulgaria's population was 7,000,039.",
        "Both the number of live births and deaths decreased in 2018.",
        "The municipality with the highest population in 2018 was Stolichna, with 1,328,120 residents."
    ]
}
\end{verbatim}

\end{tcolorbox}

\textbf{\\DOCUMENT CONTEXT} (may be truncated)

\texttt{\textcolor{blue}{\{\{document\}\}}} \\

\textbf{RESPONSE}

\end{tcolorbox}

\caption{Full prompt designed to annotate data for three tasks given a document in DocAssist: Summarization, Question Answering and Question Suggestion.
Please see the requirements for each task \texttt{\textcolor{violet}{\{\{summ\_req\}\}}}, \texttt{\textcolor{orange}{\{\{suggestion\_req\}\}}} and \texttt{\textcolor{teal}{\{\{qa\_req\}\}}} in \cref{tab:do_prompt,tab:qa_prompt,tab:sq_prompt}, respectively.
}
\label{tab:anno_prompt_full}
\end{table*}

\begin{table*}[ht]
\centering
\begin{tcolorbox}[
    colback=white,
    colframe=black,
    arc=0mm,
    boxrule=0.5pt,
    width=\textwidth
]
\small

You are an AI assistant for document analysis, performing summarization, question suggestion, and question answering. \\

\textbf{For each task:} \\
1. Analyze the given document \\
2. Determine the task (summarization, question suggestion, or question answering) \\
3. Perform the requested task \\

\textbf{Respond using this format:}
\begin{tcolorbox}[colback=gray!10,colframe=gray!10,left=1pt,right=1pt,top=1pt,bottom=1pt]
\scriptsize
\begin{verbatim}
<intent>: [summarization|question_suggestion|question_answering]
<response>
[Task-specific response here]
</response>
\end{verbatim}
\end{tcolorbox}

Now, analyze the following document and respond to the request:



\begin{tcolorbox}[colback=gray!10,colframe=gray!10,left=1pt,right=1pt,top=1pt,bottom=1pt]
\scriptsize
\begin{Verbatim}[commandchars=\\\{\}]
<document>
\textcolor{blue}{\{\{document\}\}}
</document>

<request>
\textcolor{magenta}{\{\{request\}\}}
</request>
\end{Verbatim}
\end{tcolorbox}

\end{tcolorbox}

\caption{Full prompt designed to finetune SMLs to detect and handle three tasks given a user-uploaded document in DocAssist: Summarization, Question Answering and Question Suggestion.}
\label{tab:finetuning_prompt}
\end{table*}

\begin{table*}[ht]
\centering
\caption{\textbf{Summarization task performance comparison}. SlimLM models show competitive performance: (a) SlimLM-125M outperforms SmolLM-135M-Instruct, (b) SlimLM-350M surpasses SmolLM-360M-Instruct, (c) SlimLM-450M performs comparably to Qwen2-0.5B-Instruct, and (d) SlimLM-1B approaches Qwen2-1.5B-Instruct's performance despite being smaller.}
\label{tab:main_summ}
\resizebox{\textwidth}{!}{%
\begin{tabular}{|l|c|c|c|c|c|c|c|}
\hline
\rowcolor[HTML]{A9A9A9} 
Model & BLEU $\uparrow$ & ROUGE-1 $\uparrow$ & ROUGE-2 $\uparrow$ & ROUGE-L $\uparrow$ & STS Score $\uparrow$ & GEval $\uparrow$ & Average \\
\hline
The commercial LLM & 1.00 & 1.00 & 1.00 & 1.00 & 1.00 & 0.86 & 0.9760 \\
SmolLM-135M-Instruct & 0.09 & 0.37 & 0.14 & 0.32 & 0.69 & 0.63 & 0.3762 \\
SmolLM-360M-Instruct & 0.13 & 0.42 & 0.18 & 0.36 & 0.74 & 0.71 & 0.4233 \\
Qwen2-0.5B-Instruct & 0.20 & 0.50 & 0.25 & 0.43 & 0.82 & 0.79 & 0.4985 \\
Qwen2-1.5B-Instruct & 0.26 & 0.54 & 0.31 & 0.48 & 0.84 & 0.83 & 0.5433 \\
\hline
\rowcolor[HTML]{E6E6E6}
\multicolumn{8}{|l|}{Slim Language Models (ours)} \\
\hline
\cellcolor{green!25}SlimLM-125M$^a$ & \cellcolor{green!25}0.12 & \cellcolor{green!25}0.40 & \cellcolor{green!25}0.17 & \cellcolor{green!25}0.35 & \cellcolor{green!25}0.73 & \cellcolor{green!25}0.66 & \cellcolor{green!25}0.4061 \\
SlimLM-270M & 0.17 & 0.46 & 0.22 & 0.40 & 0.79 & 0.74 & 0.4620 \\
\cellcolor{green!25}SlimLM-350M$^b$ & \cellcolor{green!25}0.16 & \cellcolor{green!25}0.45 & \cellcolor{green!25}0.22 & \cellcolor{green!25}0.39 & \cellcolor{green!25}0.78 & \cellcolor{green!25}0.74 & \cellcolor{green!25}0.4570 \\
SlimLM-450M$^c$ & 0.20 & 0.49 & 0.25 & 0.43 & 0.80 & 0.77 & 0.4893 \\
SlimLM-760M & 0.20 & 0.49 & 0.25 & 0.43 & 0.81 & 0.78 & 0.4921 \\
SlimLM-1B$^d$ & 0.23 & 0.52 & 0.28 & 0.46 & 0.82 & 0.81 & 0.5194 \\
\hline
\end{tabular}
}
\end{table*}

\begin{table*}[ht]
\centering
\caption{\textbf{Question Answering task performance comparison}. SlimLM models demonstrate strong performance: (a) SlimLM-125M outperforms SmolLM-135M-Instruct, (b) SlimLM-350M surpasses SmolLM-360M-Instruct, (c) SlimLM-450M and SlimLM-760M perform comparably to Qwen2-0.5B-Instruct, and (d) SlimLM-1B approaches Qwen2-1.5B-Instruct's performance.}
\label{tab:main_qa}
\resizebox{\textwidth}{!}{%
\begin{tabular}{|l|c|c|c|c|c|c|c|}
\hline
\rowcolor[HTML]{A9A9A9} 
Model & BLEU $\uparrow$ & ROUGE-1 $\uparrow$ & ROUGE-2 $\uparrow$ & ROUGE-L $\uparrow$ & STS Score $\uparrow$ & GEval $\uparrow$ & Average \\
\hline
The commercial LLM & 1.00 & 1.00 & 1.00 & 1.00 & 1.00 & 0.90 & 0.9830 \\
SmolLM-135M-Instruct & 0.18 & 0.45 & 0.26 & 0.42 & 0.72 & 0.56 & 0.4300 \\
SmolLM-360M-Instruct & 0.22 & 0.49 & 0.31 & 0.46 & 0.76 & 0.67 & 0.4860 \\
Qwen2-0.5B-Instruct & 0.30 & 0.57 & 0.39 & 0.54 & 0.81 & 0.79 & 0.5687 \\
Qwen2-1.5B-Instruct & 0.36 & 0.62 & 0.44 & 0.59 & 0.84 & 0.85 & 0.6157 \\
\hline
\rowcolor[HTML]{E6E6E6}
\multicolumn{8}{|l|}{Slim Language Models (ours)} \\
\hline
\cellcolor{green!25}SlimLM-125M$^a$ & \cellcolor{green!25}0.22 & \cellcolor{green!25}0.49 & \cellcolor{green!25}0.30 & \cellcolor{green!25}0.46 & \cellcolor{green!25}0.75 & \cellcolor{green!25}0.62 & \cellcolor{green!25}0.4731 \\
SlimLM-270M & 0.24 & 0.52 & 0.33 & 0.49 & 0.78 & 0.69 & 0.5077 \\
\cellcolor{green!25}SlimLM-350M$^b$ & \cellcolor{green!25}0.26 & \cellcolor{green!25}0.53 & \cellcolor{green!25}0.35 & \cellcolor{green!25}0.50 & \cellcolor{green!25}0.78 & \cellcolor{green!25}0.72 & \cellcolor{green!25}0.5246 \\
SlimLM-450M$^c$ & 0.29 & 0.56 & 0.37 & 0.53 & 0.80 & 0.75 & 0.5491 \\
SlimLM-760M$^c$ & 0.30 & 0.57 & 0.39 & 0.54 & 0.81 & 0.79 & 0.5679 \\
SlimLM-1B$^d$ & 0.32 & 0.60 & 0.41 & 0.57 & 0.83 & 0.81 & 0.5907 \\
\hline
\end{tabular}
}
\end{table*}

\begin{table*}[ht]
\centering
\caption{\textbf{Question Suggestion task performance comparison}. SlimLM models show competitive results: (a) SlimLM-125M outperforms SmolLM-135M-Instruct, (b) SlimLM-350M surpasses SmolLM-360M-Instruct, (c) SlimLM-450M and SlimLM-760M perform comparably to Qwen2-0.5B-Instruct, and (d) SlimLM-1B approaches Qwen2-1.5B-Instruct's performance in most metrics.
As Self-BLEU measures text diversity where lower scores indicate higher diversity (better), it is not included in the average scores.
}
\label{tab:main_qs}
\resizebox{\textwidth}{!}{%
\begin{tabular}{|l|c|c|c|c|c|c|c|}
\hline
\rowcolor[HTML]{A9A9A9} 
Model & BLEU $\uparrow$ & ROUGE-1 $\uparrow$ & ROUGE-2 $\uparrow$ & ROUGE-L $\uparrow$ & STS Score $\uparrow$ & Diversity $\downarrow$ & Average \\
\hline
The commercial LLM & 1.00 & 1.00 & 1.00 & 1.00 & 1.00 & 0.04 & 1.0000 \\
SmolLM-135M-Instruct & 0.04 & 0.29 & 0.11 & 0.29 & 0.49 & 0.05 & 0.2434 \\
SmolLM-360M-Instruct & 0.07 & 0.34 & 0.15 & 0.33 & 0.53 & 0.03 & 0.2837 \\
Qwen2-0.5B-Instruct & 0.12 & 0.39 & 0.20 & 0.38 & 0.59 & 0.02 & 0.3381 \\
Qwen2-1.5B-Instruct & 0.16 & 0.44 & 0.25 & 0.43 & 0.63 & 0.02 & 0.3837 \\
\hline
\rowcolor[HTML]{E6E6E6}
\multicolumn{8}{|l|}{Slim Language Models (ours)} \\
\hline
\cellcolor{green!25}SlimLM-125M$^a$ & \cellcolor{green!25}0.07 & \cellcolor{green!25}0.33 & \cellcolor{green!25}0.14 & \cellcolor{green!25}0.32 & \cellcolor{green!25}0.52 & \cellcolor{green!25}0.04 & \cellcolor{green!25}0.2754 \\
SlimLM-270M & 0.10 & 0.37 & 0.18 & 0.36 & 0.56 & 0.03 & 0.3122 \\
\cellcolor{green!25}SlimLM-350M$^b$ & \cellcolor{green!25}0.10 & \cellcolor{green!25}0.36 & \cellcolor{green!25}0.18 & \cellcolor{green!25}0.35 & \cellcolor{green!25}0.56 & \cellcolor{green!25}0.03 & \cellcolor{green!25}0.3109 \\
SlimLM-450M$^c$ & 0.11 & 0.39 & 0.20 & 0.38 & 0.59 & 0.02 & 0.3326 \\
SlimLM-760M$^c$ & 0.12 & 0.39 & 0.20 & 0.38 & 0.59 & 0.02 & 0.3389 \\
SlimLM-1B$^d$ & 0.15 & 0.43 & 0.24 & 0.42 & 0.62 & 0.02 & 0.3713 \\
\hline
\end{tabular}
}
\end{table*}

\begin{table*}[ht]
\centering
\resizebox{\textwidth}{!}{%
\begin{tabular}{|l|c|c|c|c|c|c|}
\hline
\rowcolor[HTML]{A9A9A9} 
 & \# Layers & \# Heads & Model Dimension & Learning Rate & Global Batch Size & \# Trained Tokens (billions) \\
\hline
SlimLM-125M & 12 & 12 & 2,048 & 3e-4 & 2,048 & 627 \\
\hline
SlimLM-270M & 16 & 64 & 2,048 & 4e-4 & 2,048 & 627 \\
\hline
SlimLM-350M & 24 & 16 & 2,048 & 3e-4 & 2,048 & 627 \\
\hline
SlimLM-450M & 20 & 64 & 2,048 & 3e-4 & 2,048 & 627 \\
\hline
SlimLM-760M & 24 & 12 & 2,048 & 3e-4 & 2,048 & 627 \\
\hline
SlimLM-1B & 24 & 16 & 2,048 & 2e-4 & 2,048 & 627 \\
\hline
\end{tabular}
}
\caption{Specifications of SlimLM models and hyperparameters for pre-training. 
Fine-tuning parameters are consistent across all models: learning rate of 5e-6, global batch size of 48, and 2 epochs ($\sim$725M trained tokens).}
\label{tab:slimlm_models}
\end{table*}

\end{document}